\documentclass[runningheads]{llncs}

\usepackage{amsmath} 
\usepackage{amssymb} 
\usepackage{graphicx}
\usepackage{algorithm}
\usepackage{algorithmic}
\usepackage{subfigure}
\usepackage{figsize}
\usepackage{verbatim} 
\usepackage{multirow}
\usepackage{array}

\usepackage{url}
\urldef{\mailsa}\path|{shai.haim, toby.walsh}@nicta.com.au|

\begin{document}

\mainmatter  

\title{Restart Strategy Selection using Machine Learning Techniques}

\titlerunning{Restart Strategy Selection using Machine Learning Techniques}

\author{Shai Haim\and Toby Walsh}
\authorrunning{S. Haim and T. Walsh}

\institute{NICTA and School of Computer Science and Engineering\\
University of New South Wales\\
Sydney, Australia\\
\mailsa\\}

\maketitle

\begin{abstract}

Restart strategies are an important factor in the performance of conflict-driven Davis Putnam style SAT solvers.  Selecting a good restart strategy for a problem instance can enhance the performance of a solver. Inspired by recent success applying machine learning techniques to predict the runtime of SAT solvers, we present a method which uses machine learning to boost solver performance through a smart selection of the restart strategy. Based on easy to compute features, we train both a satisfiability classifier and runtime models. We use these models to choose between restart strategies. We present experimental results comparing this technique with the most commonly used restart strategies. Our results demonstrate that machine learning is effective in improving solver performance.

\end{abstract}

\section{Introduction}

Restarts have been shown to boost the performance of backtracking SAT solvers (see for example \cite{hua:1},\cite{wu:vbe}). 
A restart strategy ($t_{1}$,$t_{2}$,$t_{3}$,...) is a sequence of restart lengths that the solver follows in the course of its execution. The solver first performs $t_{1}$ steps (in case of SAT solvers a step is usually a conflict). If a solution is not found, the solver abandons its current partial assignment and starts over. The second time it runs for $t_{2}$ steps, and so on.  Luby, Sinclair and Zuckerman \cite{lub:sin:zuc} show that for each instance there exists $t^{*}$, an optimal restart length that leads to the optimal restart strategy ($t^{*}$,$t^{*}$,$t^{*}$,...). In order to calculate $t^{*}$, one needs to have full knowledge of the runtime distribution (RTD) of the instance, a condition which is rarely met in practical cases. 

Since the RTD is not known, solvers commonly use ``Universal Restart Strategies''. These strategies do not assume prior knowledge of the RTD and they attempt to perform well on any given instance. Huang \cite{hua:1} shows that when applied with Conflict Driven Clause Learning solvers (CDCL), none of the commonly used universal strategies dominates all others on all benchmark families. He also demonstrates the great influence on the runtime of different restart strategies, when all its other parameters are fixed.

In this paper we show that the recent success in applying machine learning techniques to estimate solvers' runtimes can be harnessed to improve solvers' performance. We start by discussing the different universal strategies and recent machine learning success in Sect. \ref{Sec:background}. In Sect. \ref{Sec:restart_strategy_portfolio} we present \emph{LMPick}, a restart strategy portfolio based solver. Experimental results are presented and analyzed in Sect. \ref{Sec:results}. We conclude and suggest optional future study in Sect. \ref{Sec:conclusion}.

\section{Background}
\label{Sec:background}

Competitive DPLL solvers typically use restarts. Most use ``universal'' strategies, while some use ``dynamic'' restart schemes, that induce or delay restarts (such as the ones presented in \cite{bie:1} and \cite{ryv:str}).

Currently, the most commonly used universal strategies fall into one of the following categories:
\begin{itemize}
\item \emph{Fixed strategy - } (\cite{gom:sel:kau}). In this strategy a restart takes place every constant number of conflicts. While some solvers allow for a very short interval between restarts, others allow for longer periods, but generally fixed strategies lead to a frequent restart pattern. Examples of its use can be found in BerkMin \cite{gol:nov} (where the fixed restart size is 550 conflicts) and Seige \cite{rya} (fixed size is 16000 conflicts). 
\item \emph{Geometric strategy - } (\cite{wal}). In this strategy the size of restarts grows geometrically. This strategy is defined using an initial restart size and a geometric factor. Wu and van Beek \cite{wu:vbe} show that the expected runtime of this strategy can be unbounded worse than the optimal fixed strategy in the worst case. They also present several conditions which, if met, guarantee that the geometric strategy would yield a performance improvement. This strategy is used by MiniSat v1.14 \cite{een:sor} with initial restart size of 100 conflicts and a geometric factor of 1.5. 
\item \emph{Luby Strategy - } (\cite{lub:sin:zuc}). In this strategy the length of restart $i$ is $u \cdot t_{i}$ when $u$ is a constant ``unit size'' and 
\[
t_{i} = 
\begin{cases} 
2^{k-1}\text{, } & \text{if } i=2^{k}-1 \\
t_{i-2^{k-1}+1}\text{, } & \text{if } 2^{k-1}\leq i < 2^{k}-1
\end{cases}
\]

The first elements of this sequence are 1,1,2,1,1,2,4,1,1,2,1,1,2,4,8,1,1,... Luby, Sinclair and Zuckerman \cite{lub:sin:zuc} show that the performance of this strategy is within a logarithmic factor of the true optimal strategy, and that any universal strategy that outperforms their strategy will not do it by more than a constant factor. These results apply to pure Las Vegas algorithms, and do not immediately apply to CDCL solvers in which learnt clauses are kept across restarts. The effectiveness of the strategy in CDCL solvers appears mixed (\cite{hua:1},\cite{rua:hor:kau:2}) and there is still no theoretical work that that analyzes its effectiveness in such solvers. However, Luby's restart strategy is used by several competitive solvers including MiniSat2.1 and TiniSat.
\item \emph{Nested Restart strategy - } (\cite{bie:2}) This strategy and can be seen as a simplified version of the Luby strategy. After every iteration the restart length grows geometrically until it reaches a higher bound, at this point the restart size is reset to the initial value and the higher bound is increased geometrically. This strategy is used by PicoSat \cite{bie:2} and Barcelogic \footnote{http://www.barcelogic.org/}.
\end{itemize}

Previous work shows that restart strategies perform differently on different data sets. Huang \cite{hua:1} compares the performance of different strategies both for benchmark families and different benchmarks. He shows that there is no one strategy that outperformed all others across all benchmark families which suggests that adapting a strategy to a benchmark family, or even a single benchmark, could lead to performance gain. This suggests that choosing the best strategy from a set of strategies could improve the overall runtime, and for some benchmark families, improves it significantly.

Machine learning was previously shown to be an effective way to predict the runtime of SAT solvers. SatZilla \cite{xu:hut:hoo:ley} is a portfolio based solver that uses machine learning to predict which of the solvers it uses is optimal for a given instance. SatZilla uses an hierarchical approach \cite{xu:hoo:ley} and can use different evaluation criteria for solver performance. SatZilla utilizes runtime estimations to pick the best solver from a solver portfolio. The solvers are used as-is, and SatZilla does not have any control over their execution. SatZilla was shown to be very effective in the SAT competition of 2007. Two other Machine Learning based approaches for local search and CDCL are presented in \cite{hut:ham:hoo:ley} and \cite{bre:mit} respectively. 

Ruan et al. \cite{rua:hor:kau:1} suggest a way to use dynamic programming to derive dynamic restart strategies that are improved during search using data gathered in the beginning of the search. This idea corresponds with the ``Observation Window'' that we will discuss in the next section. There are several differences between this work and ours. One important difference is that their technique chooses a different instance from the ensemble at each restart. While our intention is to solve each of the instances in the ensemble, it seems their technique is geared towards a different goal, where the solver is given an ensemble of instances and is required to solve as many of them as possible.

Another approach for runtime estimation was presented in our previous work \cite{hai:wal}. In that paper we introduce a Linear Model Predictor (LMP) which demonstrates that runtime estimation can also be achieved using parameters that are gathered in an online manner, while a search is taking place, as opposed to the mostly static features gathered by SatZilla. Another difference between the methods is the way training instances are used. While SatZilla uses a large number of instances that are not tightly related, LMP uses a much smaller set of problems, but these should be more homogeneous.

\section{\emph{LMPick}: A Restart-Strategy Selector}
\label{Sec:restart_strategy_portfolio}

Since restart strategies are an important factor in the performance of DPLL style SAT solvers, a selection of a good restart strategy for a given instance should improve the performance of the solver for that instance. We suggest that by using supervised machine learning, it is possible to select a good restart strategy for a given instance. We present \emph{LMPick}, a machine learning based technique which enhances CDCL solvers' performance.  

\subsection{Restart Strategies Portfolio}

\emph{LMPick} uses a portfolio of restart strategies from which it chooses the best one for a given instance.
Following \cite{hua:1} we recognize several restart strategies that have shown to be effective on one benchmark family or more.
We chose 9 restart strategies that represent, to our understanding, a good mapping of commonly used restart strategies.

\begin{itemize}
\item \emph{luby-32 - } A Luby restart strategy with a ``unit run'' of 32 conflicts. This strategy represents a Luby restart strategy with a relatively small ``unit run''. This technique was shown to be effective by Huang \cite{hua:1}.
\item \emph{luby-512 - }  A Luby restart strategy with a ``unit run'' of 512 conflicts. This strategy represents a Luby restart strategy with a larger ``unit run''. This is the original restart scheme used by TiniSat, the solver we used in this study.
\item \emph{Fixed-512 - } A fixed restart scheme with a restart size of 512 conflicts. Similar restart schemes that are used by solvers are BerkMin's \emph{Fixed-550} \cite{gol:nov}, and the 2004 version of zChaff,  \emph{Fixed-700} \cite{mos:mad:zha:mal}.
\item \emph{Fixed-4096 - } A fixed balance scheme with a restart size of 4096 conflicts. We chose this restart scheme because it balances the short and long fixed schemes, and because it performed very well in our preliminary tests. 
\item \emph{Fixed-16384 - } A fixed balance scheme with a restart size of 16384 conflicts. A longer fixed strategy, similar to the one used by Siege \cite{rya} (\emph{Fixed-16000}).
\item \emph{Geometric-1.1 - } A geometric restart scheme with a first restart size of 32 conflicts and a geometric factor of 1.1.  inspired by the one used by Hunag \cite{hua:1}.
\item \emph{Geometric-1.5 - } A geometric restart scheme with a first restart size of 100 conflicts and a geometric factor of 1.5. This is the restart scheme used in MiniSat v1.14 \cite{een:sor}.
\item \emph{Nested-1.1 - } A nested restart strategy, with an inner value of 100 conflicts, an outer value of 1000 values and a geometric factor of 1.1. This strategy is parameterized as the one used by PicoSAT \cite{bie:2}.
\item \emph{Nested-1.5 - } A nested restart strategy, with an inner value of 100 conflicts, an outer value of 1000 values and a geometric factor of 1.5. Inspired by the results presented in \cite{ryv:str}.
\end{itemize}

\begin{figure}
\centering
\subtable{
{\scriptsize
\begin{tabular}{p{5.4cm}}
\small\textbf{Set I:} \\
1. \textbf{Number of variable}\\
2. \textbf{Number of clauses}\\
\\
\\
\small\textbf{Set II:} \\
3. \textbf{Number of variable}\\
4. \textbf{Number of clauses}\\
5. \textbf{Variable to clauses ratio}\\
6. \textbf{Number of binary clauses}\\
7. \textbf{Number of ternary clauses}\\
8. \textbf{Number of horn clauses}\\
9-12. \textbf{VCG - Variable nodes degree statistics}: mean, variation coefficient, min and max\\  
13-16. \textbf{VCG - Clause nodes degree statistics}: mean, variation coefficient, min and max\\
17-20. \textbf{Number of occurrences in Horn clauses}: mean, variation coefficient, min and max\\
21-24. \textbf{Ratio of positive and negative occurrences of variables}: mean, variation coefficient, min and max\\
25. \textbf{Number of assigned variable}
\\\end{tabular}}}
\subtable{
{\scriptsize
\begin{tabular}{p{6.2cm}}
\small\textbf{Set III:} \\
26-29. \textbf{Backjump Size}: mean, variation coefficient, min and max\\
30-33. \textbf{Search Depth}: mean, variation coefficient, min and max\\
34-37. \textbf{log(WBE) value} : mean, variation coefficient, min and max\\
\\
\small\textbf{Set IV:} \\
38. \textbf{Number of variable}\\
39. \textbf{Number of clauses}\\
40. \textbf{Variable to clauses ratio}\\
41. \textbf{Number of binary clauses}\\
42. \textbf{Number of ternary clauses}\\
43. \textbf{Number of horn clauses}\\
44-47. \textbf{VCG - Variable nodes degree statistics}: mean, variation coefficient, min and max\\  
48-51. \textbf{VCG - Clause nodes degree statistics}: mean, variation coefficient, min and max\\
52-55. \textbf{Number of occurrences in Horn clauses}: mean, variation coefficient, min and max\\
56-59. \textbf{Ratio of positive and negative occurrences of variables}: mean, variation coefficient, min and max\\
60. \textbf{Number of assigned variable}
\\\end{tabular}}}
\caption{A list of features used to build models.}
\label{Fig:featurvector}
\end{figure}

\subsection{Supervised Machine Learning}

Satsifiable and unsatisfiable instances from the same benchmark family tend to have different runtime distributions \cite{fro:ris}. A runtime prediction model that is trained using both SAT and UNSAT instances performs worse than a homogeneous model. It is better to train a layer of two models, one trained with satisfiable instances ($M_{sat}$) and the other with unsatisfiable instances ($M_{unsat}$). Since in most cases we do not know whether a given instance is satisfiable or not we need to determine which of the models is the correct one to query for a given instance according to its probability to be satisfiable. Previous work (\cite{xu:hoo:ley},\cite{dev:osu}) suggests that machine learning can be successfully used for this task as well. A classifier can be trained to estimate the probability of an instance to be satisfiable. Some classification techniques perform better than others, but it seems that for most benchmark families, a classifier with 80\% accuracy or more is achievable.

Using supervised machine learning, we train models offline in order to use them for predictions online. For every training 
example $t \in \cal T$, where $\cal T$ is the training set, we gather the feature vector $x=\left\{{x_{1},x_{2},\ldots,x_{n}}\right\}$ using the features presented in section \ref{sec:featurevector}. Once the raw data is gathered, we perform a feature selection. We repeatedly remove the feature with the smallest standardised coefficient until no improvement is observed based on the standard AIC (Akaike Information Criterion). We then searched and eliminate co-linear features in the chosen set. The reduced feature vector $\hat{x}$ is then used to train a classifier and several runtime prediction models. The classifier predicts the probability of an instance to be satisfiable and the runtime models predict cpu-runtime. \emph{LMPick} trains one classifier, but two runtime models for each restart strategy $s \in S$ (where $S$ is the set of all participating strategies) to the total of $2|S|$ models. Each training instance is used to train the satisfiability classifier, labeled with its satisfiability class, and $|S|$ runtime models, for each model it is labeled with the appropriate runtime.

As the classifier, we used a Logistic Regression technique. Any classifier that returns probabilities would be suitable. We found Logistic Regression to be a simple yet effective classifier which was also robust enough to deal with different data sets. We have considered both Sparse Multinomial Linear Regression \cite{kri:fig:car:har} (suggested to be effective for this task in \cite{xu:hut:hoo:ley}), and the classifiers suggested by Devlin and O'Sullivan in \cite{dev:osu}, but the result of all classifiers were on par when using the presented feature vector on our datasets.

For the runtime prediction models we used Ridge Linear Regression. Using ridge linear regression, we fit our coefficient vector $w$ to create a linear predictor $f_{w}\left(\hat{x}\right) =w^{T}\hat{x}_{i}$. We chose ridge regression, since it is a quick and simple technique for numerical prediction, and it was shown to be effective in the Linear Model Predictor (LMP) \cite{hai:wal}. While LMP predicts the log of number of conflicts, in  this work we found that predicting cpu-runtime is more effective as a selection criterion for restart strategies. Using the number of conflicts as a selection criterion tends to bias the selection towards frequent restart strategies for large instances. This is because an instance with many variables spends more time going down the first branch to a conflict after a restart. This work is unaccounted for when conflicts are used as the cost criterion. Hence a very frequent restart strategy might be very effective in the number of conflict while much less effective in cpu-time. 

\subsection{Feature Vector}
\label{sec:featurevector}
There are 4 different sets of features that we used in this study, all are inspired by the two previously discussed techniques - SatZilla \cite{xu:hut:hoo:ley} and LMP \cite{hai:wal}. The first set include only the number of variables and the number of clauses in the original clause database. These values are the only ones that are not normalized. The second set includes variables that are gathered before the solver starts but after removing clauses that are already satisfiable, shrinking clauses with multiple appearances and propagating unit clauses in the original formula. These features are all normalized appropriately. They are inspired by SatZilla and were first suggested in \cite{nud:ley:hoo:dev:sho}. The third set include statistics that are gathered during the ``Observation Window'', this is a period where we analyze the behavior of the solver while solving the instance. The ``Observation Window'' was first used in \cite{hai:wal}. The way the observation window is used in this study will be discussed shortly. The variables in this set are the only ones which are DPLL dependent. The last set includes the same features as the second, but they are calculated at the end of the observation window. A full list of the features is presented in Fig. \ref{Fig:featurvector}. For further explanation about these features see \cite{nud:ley:hoo:dev:sho} and \cite{hai:wal}.

\begin{figure}
\centering
\label{Fig:steps}
\includegraphics[width=8.20cm]{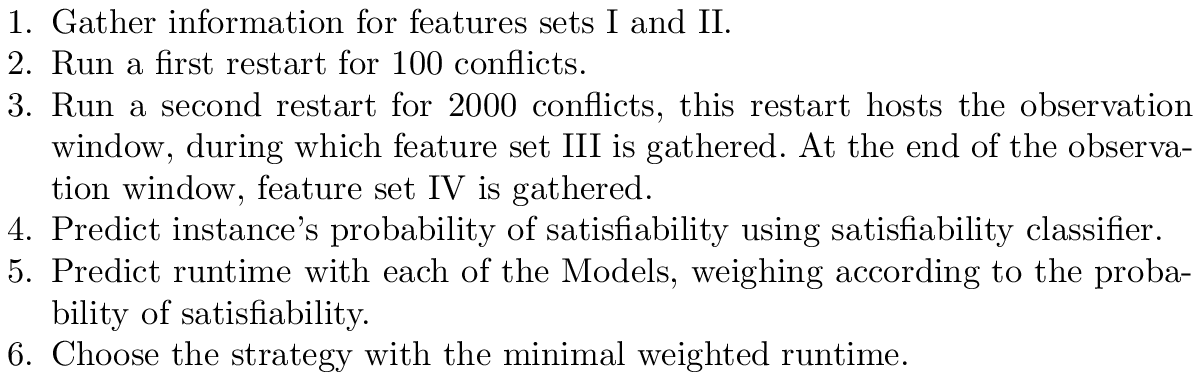}
\caption{Steps in the operation of a restart strategy portfolio based solver. Features sets I through IV are presented in Fig. \ref{Fig:featurvector}.}
\end{figure}

\subsection{Operation of the Solver}
Once all runtime models are fitted and the satisfiability classifier is trained, we can use them to improve performance for future instances.
The steps that are taken by \emph{LMPick} are presented in Fig. \ref{Fig:steps}. 
 
Since no prediction can be made before the observation window is terminated, and since we favor an early estimation, it is important that the observation window should terminate early in the search. In our preliminary testings we have noticed that the first restart tends to be very noisy, and that results are better if data is collected in the second restart onwards. We have tried several options for the observation window location and size, eventually we opted for a first restart which is very short (100 conflicts), followed by a second restart (of size 2000) which hosts the observation window. Hence the observation window is closed and all data is gathered after 2100 conflicts. 

Once the feature vector $x$ is gathered it is used with the classifier to determine the probability of the instance to be satisfiable, $P(sat | x)$. For each of the strategies, both models are queried and a best strategy $s_{b}(x)$ is picked using

\begin{displaymath}
s_{b}(x) = \operatorname*{arg\,min}_{s \in S} [ P(sat | x) \cdot M_{sat,s}(x) + P(unsat | x) \cdot M_{unsat,s}(x)].
\end{displaymath}

The restart strategy which is predicted to be the first to terminate is picked, and the solver starts following this strategy from the next restart onwards.
Although restart strategies are usually followed from the beginning of the search, we do not want to lose the learned clauses from the first 2100 conflicts. Therefore, we continue the current solving process and keep the already learnt clauses. We denote the restart sequence that takes place from the first restart to termination as $LMPick_{s_{b}}$. It is important to note that $s_{b} \neq LMPick_{s_{b}}$.

\begin{figure}
\centering
\subfigure[$sat$] 
{
    \label{Fig:script:bmc:sat}
    \includegraphics[width=5.20cm]{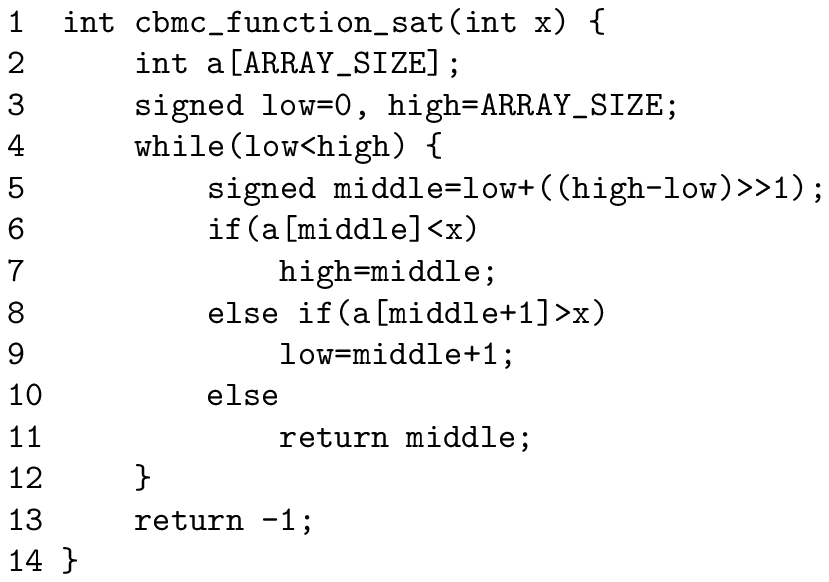}
}
\hspace{0.1cm}
\subfigure[$unsat$] 
{
    \label{Fig:script:bmc:unsat}
    \includegraphics[width=5.20cm]{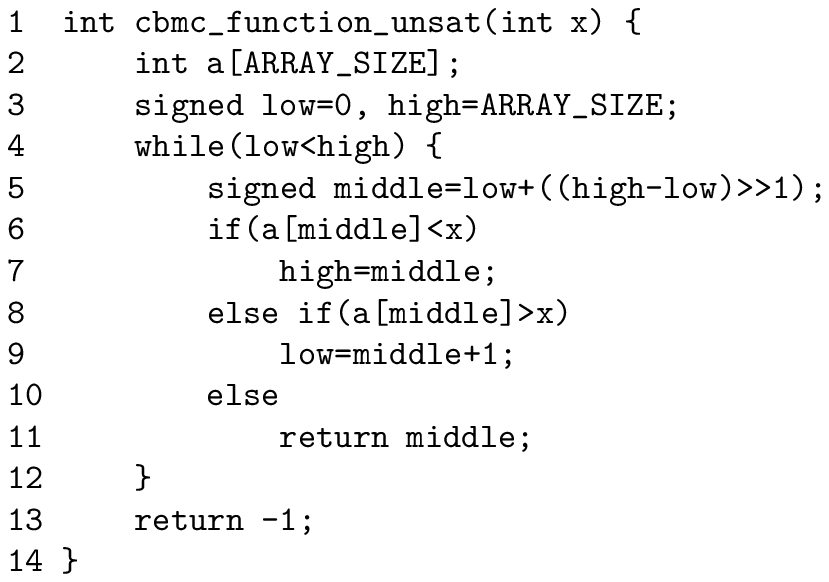}
}
\caption{Code verified by CBMC to generate the $bmc$ dataset. Different instances are made using different ARRAY\_SIZE values and a different number of unwinding iterations.}
\label{Fig:script:bmc} 
\end{figure} 

\section {Results}
\label{Sec:results}
\subsection{Experiment Settings}
In this study we used TiniSat (version 0.22) \cite{hua:2} as the basic solver. TiniSat is a lightweight DPLL style solver that was first presented in the SAT Race of 2006. TiniSat is a modern solver that uses clause learning and a unique decision heuristic that generally favours variables from recent assignments (as in BerkMin \cite{gol:nov}) and uses VSIDS \cite{mos:mad:zha:mal} over the literals as a backup. We chose to use TiniSat since (i) it is tuned in a way that would make comparison of restart strategies more meaningful \cite{hua:1} and (ii) it is a compact and straightforward implementation which allows for greater ease of use. TiniSat is not equipped with a pre-processor, and we have not used any in our study. By default, TiniSat uses a Luby restart strategy with a run unit of 512 conflicts.
All our experiments were conducted on a cluster of 14 Dual Intel Xeon CPUs with EM64T (64-bit) extensions, running at 3.2GHz with 4GB of RAM under Debian GNU/Linux 4.0. By implementing a runtime cutoff of 90 minutes per instance, we managed to complete all experiments in approximately 290 CPU days.

\begin{table}
	\centering
	\caption{Summary of features of datasets. For each dataset the following details are presented: The dataset's classification (Class), the number of instances (Ins.) and its size in MB (all file are zipped). Also, we present the time (in hours) it took for all 9 restart strategies to solve these datasets. We present the mean time (Mean), the standard deviation (SD) and the minimal and maximal time. Runtime cutoff is 5400 seconds and it is the maximal runtime per instance.}
	\label{Table:datasets}
  \begin{tabular}{ | l | l | l | l || c | c | c | c ||}
  \hline
			
		\multirow{2}{*}{Name} & \multirow{2}{*}{Class} & \multirow{2}{*}{Ins.}   & Size  & \multicolumn{4}{c ||}{Runtime}\\
				& & & (MB) & Mean & SD & Min & Max\\\hline \hline										
		\multirow{2}{*}{\emph{bmc}}    & sat & 234 & 4,420.0 & 154.42 & 26.41 & 127.93 & 213.27\\ 
                            & unsat & 237 & 1,951.7 & 113.82 & 18.59 & 93.48 & 155.04\\\hline	
		\multirow{2}{*}{\emph{velev}}  & sat & 72 & 1,866.2 & 32.68 & 3.74 & 24.77 & 37.87\\ 
                            & unsat & 105 & 953.4 & 70.74 & 1.64 & 68.16 & 73.29\\\hline	
		\multirow{2}{*}{\emph{crypto}} & sat & 139 & 2.7 & 140.71 & 11.54 & 122.52 & 164.78\\ 
                            & unsat & 300 & 5.5 & 14.19 & 1.53 & 11.61 & 16.65\\\hline
    \multirow{2}{*}{\emph{rand}}   & sat & 457 &  1.90 & 76.93 & 14.98 & 54.00 & 105.10\\ 
                            & unsat & 601 & 2.3 & 84.82 & 10.82 & 70.87 & 103.56\\\hline		                           		
  \end{tabular}
\end{table}

\subsection{Benchmarks}
\label{results:benchmarks}
In this study we used four different distributions of SAT instances. Instances in each data set are of various difficulty. We have omitted very easy instances that are solved before the ``observation window'' terminates.
\begin{itemize}
\item \emph{bmc:} An ensemble of software verification problems generated 
using CBMC\footnote{http://www.cprover.org/cbmc/} verifying the C functions presented in Fig. \ref{Fig:script:bmc}. These two functions are almost identical, apart for a change in line 8, which causes the $sat$ script to overflow. The different instances use different array sizes and different number of unwindings. This dataset represents an ensemble of problems that are very similar and generated by the same process. We use 234 satisfiable and 237 unsatisfiable problems.
\item \emph{velev:} An ensemble of hardware formal verification problems distributed by Miroslav Velev\footnote{http://www.miroslav-velev.com/sat\_benchmarks.html. We use the following benchmark families: vliw\_sat\_2.0, vliw\_sat\_2.1, vliw\_sat\_4.0, vliw\_unsat\_2.0, vliw\_unsat\_3.0, vliw\_unsat\_4.0, pipe\_sat\_1.0, pipe\_sat\_1.1, pipe\_unsat\_1.0, pipe\_unsat\_1.1, liveness\_s\-at\_\-1.0, liveness\_\-unsat\_1.0, liveness\_\-unsat\_2.0, dlx\_iq\_\-unsat\_1.0, dlx\_iq\_\-unsat\_2.0, engine\_\-unsat\_1.0, fvp\_\-sat\_3.0, fvp\_\-unsat\_1.0, fvp\_\-unsat\_2.0, fvp\_unsat\_3.0.}. These are well studied verification hardware benchmarks. This ensemble is not as homogeneous as \emph{bmc} because it is a union of many small benchmark families. We use 72 satisfiable and 105 unsatisfiable instances.
\item \emph{crypto:} An ensemble of problems that are generated as part of an attack on the Bivium stream cipher, presented by Eibach, Pilz and V\"olkel \cite{eib:pil:vol}. This ensemble presents some interesting characteristics. While it is generated by a non-random process, the instances are significantly smaller than common industrial instances. The satisfiable instances we use were generated with 35 guesses, the unsatisfiable ones were generated with 40 guesses. The reason for this discrepancy is that unsatisfiable instances are harder to solve in this benchmark family, and different number of guesses renders the datasets too easy or too hard. We use 139 sat and 300 unsat instances.
\item \emph{rand:} An ensemble of 457 satisfiable and 601 unsatisfiable randomly generated 3-SAT problems
with 250 to 450 variables and a clause-to-var ratio of 4.1 to 5.0.  
\end{itemize}

Some further data about these data sets is presented in Table \ref{Table:datasets}. We would like to draw the reader's attention to the ``SD'' column. This column presents the standard deviation observed when the problem is solved by all 9 restart strategies. A small value indicates that all restart strategies perform on par on this data set, while a large value indicates that runtimes are scattered.

Each data set is split into training and testing sets. Instances in the training set are used to train $M_{sat}$, $M_{unsat}$ and the classifier while the instances in the testing set are only used to generate the results. We used a \emph{10-fold} cross validation technique after randomly shuffling all instances.

\subsection{Restart Strategy Portfolio Performance}
Table \ref{Table:runtimes} demonstrates the effectiveness of \emph{LMPick}. We present the performance of each of the 9 restart strategies on each of the data sets. We use two matrices: The number of instances solved within the cutoff time of 90 minutes, and the total time it took to solved the entire data set. Timed out instances are counted as contributing 90 minutes to the total runtime. We then present the average performance achieved by all strategies. This solver (denoted \emph{Random-pick}) represents the expected behavior when there is no prior knowledge regarding which of the strategies is most suitable for a given instance. The last row presents the results we get using the \emph{LMPick} process. 

This table shows that for all of the data sets \emph{LMPick} performs better than using a randomly picked strategy, both for problems solved and for total runtime. In terms of problem solved, \emph{LMPick} performs better than any given restart strategy in two of the cases, and performs on par with the optimal strategy in two other cases. The ``Total'' column shows that for these data sets, \emph{LMPick} would solve more instances in less time than any single restart strategy.

\def\sattable{\scriptsize{SAT}}
\def\unsattable{\scriptsize{UNSAT}}

\begin{table}
	\centering
	\caption{Performance comparison of all strategies over data sets, two metrics (M) are presented: number of solved instances (S) and total runtime (T, in seconds).  Average solver behaviour (a randomly picked solver) is presented as \emph{Rand.}. Runtime cutoff is 5400 second. Unsolved instances are considered to contribute 5400 seconds to the total runtime.}
	\label{Table:runtimes}
  \begin{tabular}{ | l | l || c | c || c | c || c | c || c | c || c |}
  \hline
		
		\multirow{2}{*}{$Strategy$} & \multirow{2}{*}{$M$} & \multicolumn{2}{c ||}{$bmc$} & \multicolumn{2}{c ||}{$velev$} & \multicolumn{2}{c ||}{$crypto$} & \multicolumn{2}{c ||}{$random$} & \multirow{2}{*}{$Total$} \\
		&  & \sattable & \unsattable & \sattable & \unsattable & \sattable & \unsattable & \sattable & \unsattable & \\\hline \hline
		\multirow{2}{*}{\emph{Luby-32}} & S & 192 & 220 & 53& 64 & 67 & 300 & 427 & 561 & 1884 \\
		                           & T   & 143.26 & 98.68 & 37.86 & 72.30 & 138.02 & 16.65 & 90.34 & 96.60 & 693.74 \\\hline
		\multirow{2}{*}{\emph{Luby-512}} & S & 196 & \textbf{226} & 59 & 64 & 71 & 300 & 431 & 564 & 1911 \\
		                           & T   & 132.90 & \textbf{93.48} & 31.72 & 71.48 & 141.19 & 14.30 & 74.89 & 86.44 & 646.38 \\\hline
		\multirow{2}{*}{\emph{Fix-512}} & S & 142 & 187 & 58 & 62 & 50 & 300 & 414 & 558 & 1771 \\
		                           & T   & 185.73 & 134.17 & 32.66 & 73.29 & 164.78 & 15.94 & 105.10 & 103.56 & 815.24 \\\hline
		\multirow{2}{*}{\emph{Fix-4096}} & S & 199 & 213 & 59 & 64 & 73 & 300 & 431 & 563 & 1902\\
		                           & T   & 138.67 & 110.52 & 33.02 & 69.80 & 136.46 & 13.57 & 72.63 & 83.30 & 657.97 \\\hline
		\multirow{2}{*}{\emph{Fix-16384}} & S & 191 & 212 & \textbf{61} & 65 & 69 & 300 & 432 & 569  & 1899\\
		                           & T   & 157.68 & 119.88 & \textbf{24.77} & \textbf{68.16} & 147.86 & 13.10 & 65.37 & 72.34 & 669.17 \\\hline
		\multirow{2}{*}{\emph{Geom-1.1}} & S & 189 & 213 & 59 & 64 & 62 & 300 & 432 & 571 & 1890 \\
		                           & T  & 127.93 & 104.83 & 29.94 & 69.02 & 149.69 & 13.96 & 68.18 & 79.44 & 642.99\\\hline	
		\multirow{2}{*}{\emph{Geom-1.5}} & S & 128 & 174 & 54 & 62 & 84 & 300 & \textbf{439} & 573  & 1814\\
		                           & T  & 213.27 & 155.04 & 37.81 & 72.62 & 130.86 & 12.93 & \textbf{54.00} & \textbf{70.87} & 747.39 \\\hline
		\multirow{2}{*}{\emph{Nest-1.1}} & S & 192 & 218 & \textbf{61} & 65 & 76 & 300 & 423 & 566  & 1901\\
		                           & T   & 137.64 & 107.69 & 32.49 & 69.76 & 134.94 & 15.65 & 92.27 & 94.87 & 685.31\\\hline                     
 		\multirow{2}{*}{\emph{Nest-1.5}} & S & 179 & 215 & 58 & 63 & \textbf{88} & 300 & 434 & \textbf{574} & 1911 \\
		                           & T   & 152.67 & 100.12 & 33.82 & 70.29 & \textbf{122.52} & \textbf{11.61} & 69.60 & 75.93 & 636.56 \\\hline \hline 
 		\multirow{2}{*}{\textbf{\textbf{\emph{Rand.}}}} & S & 178.67 & 208.67 & 58 & 63.66 & 62.77 & 300 & 429.22 & 566.56 & 1862.04 \\
		                          & T  & 154.42 & 113.82 & 32.68 & 70.74 & 140.71 & 14.19 & 76.93 & 84.82 & 666.98\\\hline \hline
		\multirow{2}{*}{\textbf{\emph{\textbf{LMPick}}}} & S & \textbf{203} & 221 & \textbf{61} & \textbf{66} & 72 & 300 & 435 & 571  & \textbf{1929}\\
		                           & T  & \textbf{124.57} & 97.53 & 28.09 & 68.98 & 138.17 & 12.02 & 68.00 & 80.86 & \textbf{618.23} \\\hline 
				                           		                                              		                           		
  \end{tabular}
\end{table}

For the \emph{bmc} data set, \emph{LMPick} performs significantly better than any other restart scheme for satisfiable instances. It solves 4 instances more than the best strategy (\emph{Static-4096}), and the total runtime shows 19.33\% improvement over the average \emph{Random-pick} strategy. For unsatisfiable instances, it performs worse. Although it is second amongst the strategies, it is clearly worse than \emph{Luby-512} that solves 6 more instances. Comparing the standard deviation of these two data sets in Table \ref{Table:datasets} does not explain this descripency in performance. Both show large, almost similar, coefficients (sat's standard deviation is 154.42, and its variation coefficient is 0.171 while unsat's standard deviation is 113.82 and its variation coefficient is 0.1633), this indicates a high potential of improvement for both data sets. We conjecture that the reason for the poorer performance lies in the fact that both the classifier, and the runtime models are more accurate for \emph{bmc-sat} than for \emph{bmc-unsat}.

\begin{table}
	\centering
	\caption{Accuracy results for the satisfiability classifier. Figures represent the percent of correctly classified instances.}
	\label{Table:classifier}  \begin{tabular}{ | l || c | c | c | c ||}
  \hline
		Class 		& $bmc$ & $velev$ & $crypto$ &  $rand$ \\\hline				
    SAT 		  & 100.00\% & 97.22\% & 99.28\%  & 82.49\% \\ 
    UNSAT			& 83.54\%  & 82.85\% & 100.00\% & 94.51\% \\
    ALL 			& 91.72\%  & 88.70\% & 99.77\%  & 89.31\% \\\hline
                            
		\hline
  \end{tabular}
\end{table}

Table \ref{Table:classifier} presents the performance of the satisfiability classifier. The figures in the table represent the percent of instances that were predicted correctly. It is important to note that since we did not apply a ``winner takes all'' approach, in some of the cases where the classification was correct but not with 100\% certainty, the wrong model was also considered. We can see that for \emph{bmc}, satisfiable instances are classifed correctly every time, while unsatisfiable ones are classified correctly only 83.54\% of the times. In order to check the influence of these classifaction errors on the performance of \emph{LMPick} for this dataset, we ran another experiment, where we used a classification oracle. \emph{LMPick}'s performance is improved, and it solved 224 instances, which are 3 more than the original results. This shows the effect of a good classification technique over the overall performance.

\begin{table}
	\centering
	\caption{Performance of \emph{LMPick}'s chosen restart strategy ($s_{b}$) in comparison with other strategies. The figures represent the percent of strategies that outperform $s_{b}$ for a given data set.}
	\label{Table:bestpick_location}  
	\begin{tabular}{ | l || c | c | c | c ||}
  \hline
		Class 		& $bmc$ & $velev$ & $crypto$ &  $rand$ \\\hline			
    SAT 		  & 30.18\%  & 35.58\%  & 32.24\%  & 39.29\%\\ 
    UNSAT			& 32.01\%  & 37.05\%  & 14.77\%  & 39.61\%\\
                            
		\hline
  \end{tabular}
\end{table}

In order for \emph{LMPick} to enhance the solver's performance, runtime estimation models need to perform well. Nevertheless, it is not crucial that each model's prediction is accurate. The important factor is the relative order of these predictions. We would prefer the chosen strategy ($s_{b}$) to be amongst the best strategies for that instance. Table \ref{Table:bestpick_location} demonstrates the quality of the chosen  strategy $s_{b}$. The table presents the percentage of strategies that outperform $s_{b}$. This table shows that in all cases $s_{b}$ is better than picking a random strategy, and that for the \emph{crypto-unsat} instances it performs very well.

The difference in the performance on the \emph{crypto-sat} data set is not easily explained by the data presented so far. While both the classifier and runtime prediction models perform better than on the \emph{velev-sat}, the overall performance is worse. We conjecture that the cause of this difference is the differing likelihood of an instance being solved by multiple strategies.

\begin{figure}
\centering
\subfigure[] 
{
    \label{Fig:compare-sat-datasets:bars}
    \includegraphics[width=4.0cm, angle=270]{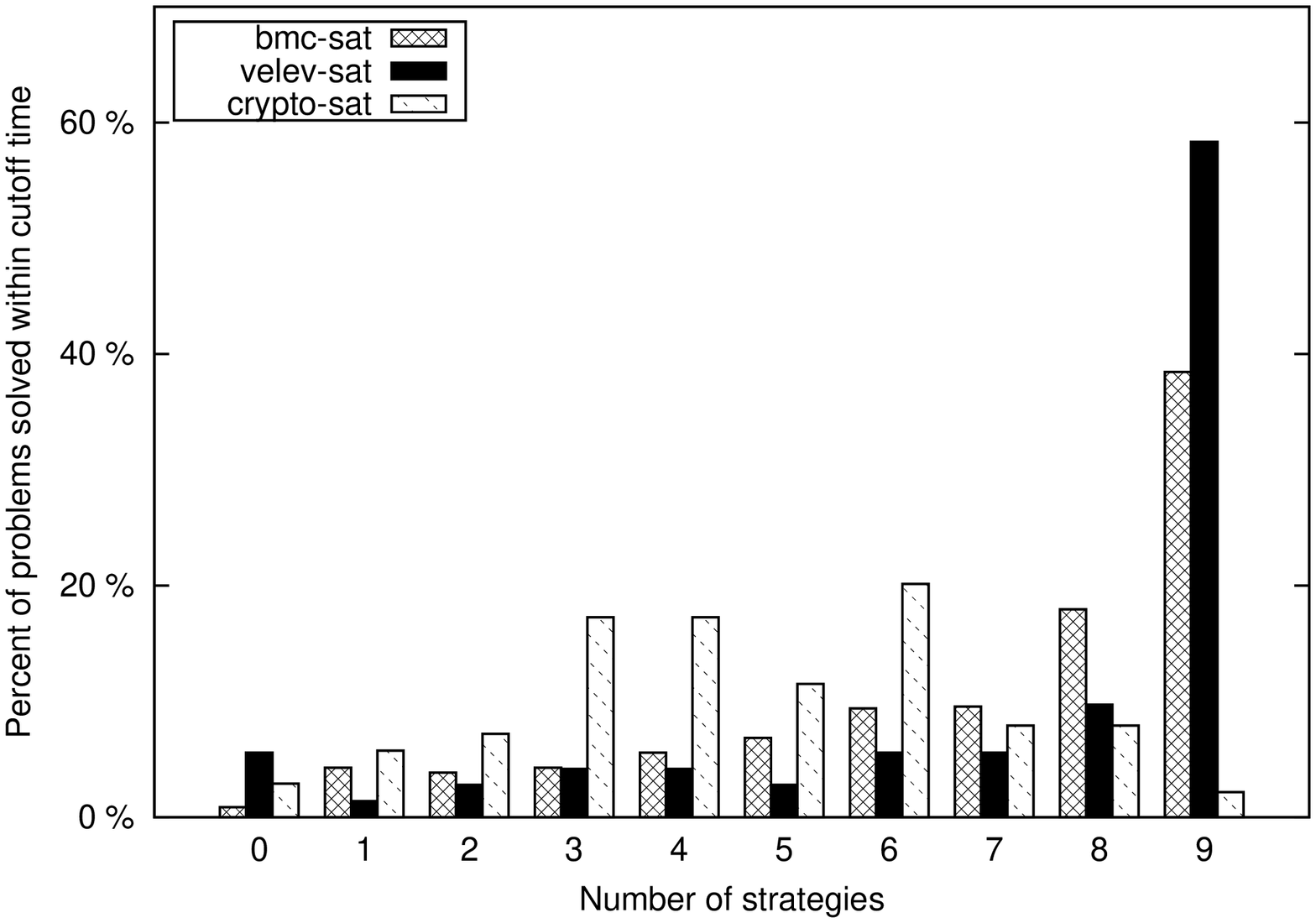}
}
\hspace{0.1cm}
\subfigure[] 
{
    \label{Fig:compare-sat-datasets:plots}
    \includegraphics[width=4.0cm, angle=270]{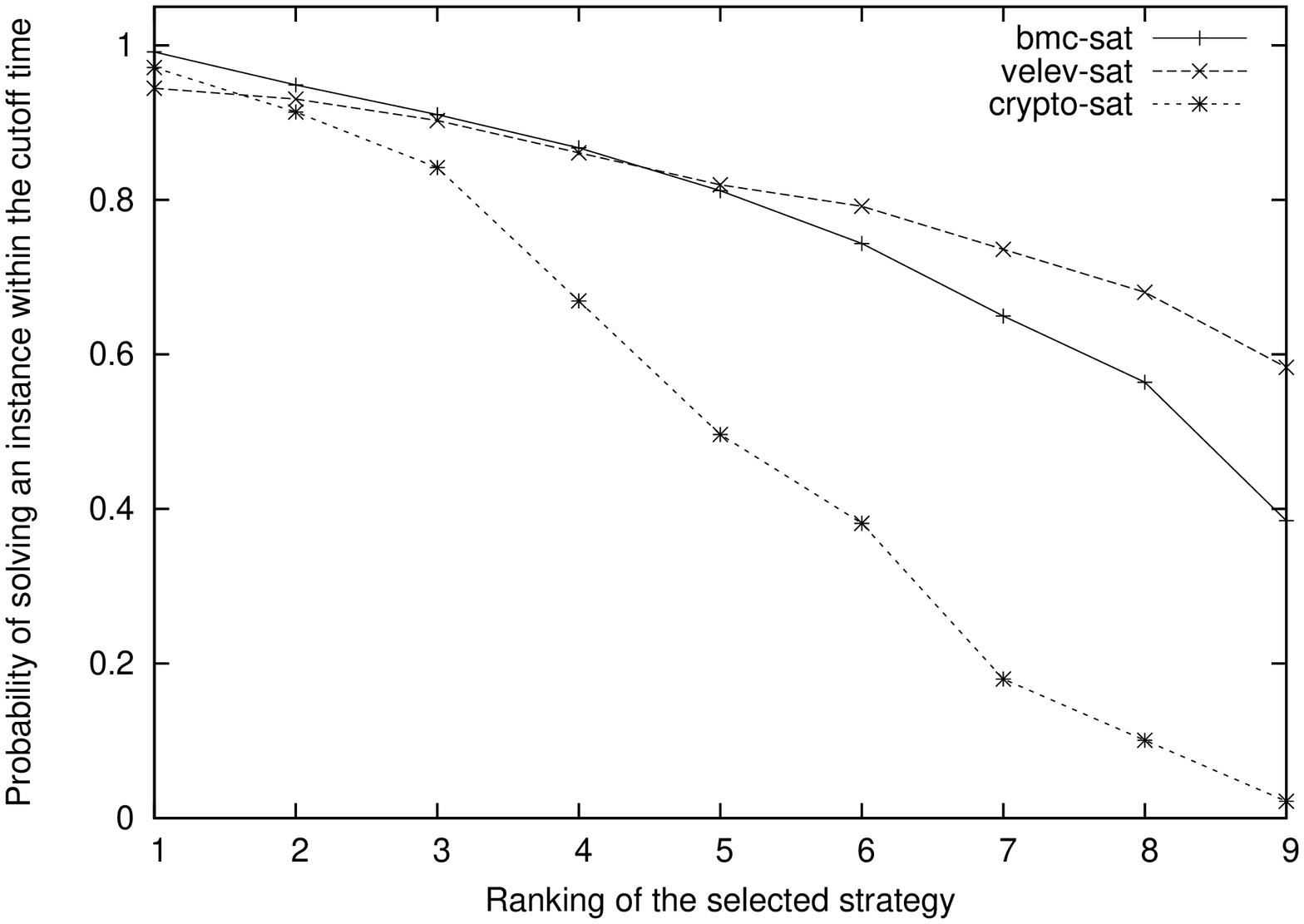}
}
\caption{A comparison of strategies' performance on three satisfiable data sets. On the left, bars represent the percent of instances that are solved (within cutoff time) by each number of strategies, when 0 means all instances that cannot be solved by any of the strategies and 9 means all instances that are solved by all strategies. On the right the plot shows the probability of solving a randomly picked instance (within cutoff time) as a function of the quality of the selection decision, when 1 means the strategy selected was the best one and 9 means it was the worst.}
\label{Fig:compare-sat-datasets}
\end{figure}

Figure \ref{Fig:compare-sat-datasets} compares the 3 non-random satisfiable data sets. In Fig. \ref{Fig:compare-sat-datasets:bars} bars represent the percent of instances in the data set that were solved within the cutoff time by each number of strategies. There is a clear difference between \emph{crypto-sat} and the other 2 data sets. While for both industrial verification based data sets, most instances were solved by the majority of strategies, for \emph{crypto-sat} many of the instances are solved by a small set of strategies. The effect of this difference is demonstrated in Fig. \ref{Fig:compare-sat-datasets:plots}. This plot presents the probability of a randomly picked instance being solved within the cutoff time given the quality of selected strategy. From left to right, the picked strategy shift from best to worse. If \emph{LMPick} picks one of the the two best strategies, the probability of all three data sets is quite similar, but when the chosen strategy is 3rd or 4th, the probability of solving a \emph{crypto-sat} instance drops significantly compared to the other two. Many instances in the \emph{crypto-sat} data set are only solved within the cutoff by a small subset of strategies, making this data set harder as a sub-optimal selection is likely to lead to a timeout.

\section{Conclusions and Future Work}
\label{Sec:conclusion}

Restart strategies have an important role in the success of DPLL style SAT solvers. The performance of different strategies varies over different benchmark families. We harness machine learning to enhance the performance of SAT solvers. We have presented \emph{LMPick}, a technique that uses both satisfiability classification and solver runtime estimation to pick promising restart strategies for instances. We have demonstrated the effectiveness of \emph{LMPick} and compared its results with the most commonly used restart strategies. We have established that in many cases \emph{LMPick} outperforms any single restart strategy and that it is never worse than a randomly picked strategy. We have also discussed the influence of different components of \emph{LMPick} on its performance.

While universal restart strategies are more commonly used than dynamic ones, dynamic strategies are getting more attention lately. An interesting continuation to this work would be to use machine learning to develop a fully dynamic restart strategy.  Such a strategy could use unsupervised machine learning algorithm to develop a dynamic restart policy for a benchmark family of problems. 
Restart strategies are not the only aspect of SAT solving influencing performance on different benchmark families. Machine learning can be also used to tune other parameters that govern the behavior of modern DPLL based solvers, namely, parameters that are commonly set manually as a result of a trial and error process, such as decision heuristic parameters, clause deletion policy, etc.

\section*{Acknowledgements}

NICTA is funded by the Department of Broadband, Communications
and the Digial Economy, and the ARC through Backing Australia's Ability
and the ICT Centre of Excellence program.

\end{document}